\documentclass[runningheads]{llncs}
\usepackage{graphicx}
\usepackage{array}
\usepackage{multirow}
\usepackage{subfigure}
\usepackage{stmaryrd}
\usepackage{mathabx}
\begin{document}
\title{How many labeled license plates are needed?}
%
%
\author{Changhao Wu\inst{1}\and
Shugong Xu\inst{1}\thanks{Corresponding author. Shanghai Institute for Advanced Communication and Data Science, Shanghai University, Shanghai, China(email: shugong@shu.edu.cn).} \and
Guocong Song\inst{2} \and
Shunqing Zhang\inst{1}}


\institute{Shanghai Institute for Advanced Communication and Data Science\\
Shanghai University, Shanghai, 200444, China\\
\email{\{wuchanghao, shugong, shunqing\}@shu.edu.cn}
\and{Playground Global \\
\email{guocong@playground.global}}}
%
\maketitle              
\pagestyle{empty}  
\thispagestyle{empty} 
\begin{abstract}
Training a good deep learning model often requires a lot of annotated data. As a large amount of labeled data is typically difficult to collect and even more difficult to annotate, data augmentation and data generation are widely used in the process of training deep neural networks. However, there is no clear common understanding on how much labeled data is needed to get satisfactory performance. In this paper, we try to address such a question using vehicle license plate character recognition as an example application. We apply computer graphic scripts and Generative Adversarial Networks to generate and augment a large number of annotated, synthesized license plate images with realistic colors, fonts, and character composition from a small number of real, manually labeled license plate images. Generated and augmented data are mixed and used as training data for the license plate recognition network modified from DenseNet. The experimental results show that the model trained from the generated mixed training data has good generalization ability, and the proposed approach achieves a new state-of-the-art accuracy on Dataset-1 and AOLP, even with a very limited number of original real license plates. In addition, the accuracy improvement caused by data generation becomes more significant when the number of labeled images is reduced. Data augmentation also plays a more significant role when the number of labeled images is increased.

\end{abstract}
\section{Introduction}
License plate recognition is one of the most important components of modern intelligent transportation systems. It has attracted the attention of many researchers. However, most existing algorithms\cite{Gou2016Vehicle,li2016reading,li2018toward,wang2017adversarial} can only work normally under certain conditions. For example, some recognition systems require sophisticated hardware to shoot high-quality images, while other systems require the vehicle to slowly pass through a fixed access opening or even stop. Accurately detecting license plates and recognizing characters in an open environment is a challenging task. The main difficulties are different license plate fonts and colors, character distortion caused by the image capture process and non-uniform illumination, and low-quality images caused by occlusion or motion blur.

\begin{figure}[t]
\centering
\includegraphics[width=0.8\textwidth]{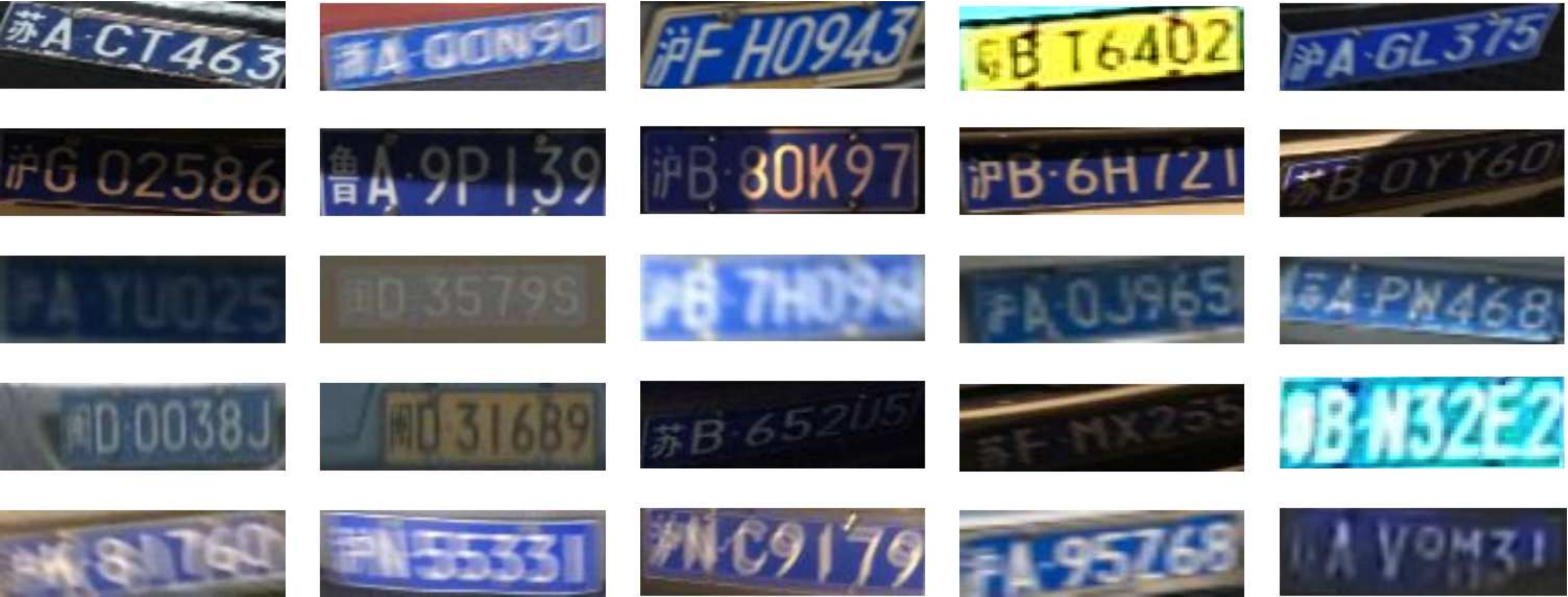}
\caption{The complex license plates images.} \label{fig:complex}
\end{figure}

In this paper, we propose a license plate recognition system, in which we cope with challenge such as, low light, low resolution, motion blur, and other harsh conditions. Fig. \ref{fig:complex} shows the license plates which can be correctly recognized by our proposed method. From top to bottom are the license plate images affected by the shooting angle, uneven illumination, low resolution, detection error and motion blur.

In general, supervised learning requires a large amount of labeled data in order to achieve good results. However, real data is not easy to obtain, the acquisition process is slow, and the data needs to be processed and annotated before it can be used for training. To achieve a higher accuracy of the annotation, manual inspection is also required.


However, the acquisition of a large amount of real data and manual annotations is very expensive. Therefore, data generation is very important for the training of license plate recognition network. We believe that the information contained in a small number of real license plates is sufficient to recognize most of the existing license plate images. However, there is no clear common understanding on how much labeled data is needed to get satisfactory performance. In this paper, we try to address such a question in vehicle license plate character recognition.

The main contributions of this paper can be summarized as the following three points:

1.We propose various methods of data generation and data augmentation. As long as we have a few labeled license plate images, a large amount of generated data can be created. We can achieve and even exceed the recognition accuracy and results of systems trained only on real images.

2.We compare the performance of various data generation and data augmentation methods to find that both data generation methods and data augmentation methods can significantly improve license plate recognition accuracy. Data augmentation plays a larger role in accuracy improvement when there are many labeled license plates but when the number of labelled license plates is small, data generation more significantly increases accuracy.

3.We apply a network that is modified from DenseNet to license plate recognition to reduce network parameters and inference time and improve accuracy.

The rest of paper is arranged as follows. In Section 2 we review the related works briefly. In Section 3 we describe the details of networks used in our approach. Experimental results are provided in Section 4, and conclusions are drawn in Section 5.

\section{Related Work}
The section introduces previous work on license plate recognition and GANs.

\subsection{License Plate Recognition}
Existing license plate recognition systems are either text segmentation-based \cite{Gou2016Vehicle,Guo2008License}, or non-segmentation-based \cite{li2016reading}. Methods that depend on segmentation first preprocess the license plate image and then segment individual characters through image processing. After this, each character is classified by a convolutional neural network. This method is very dependent on the accuracy of text segmentation, and the recognition speed is slower. A recognition method that does not require segmentation is proposed by Li et al. \cite{li2016reading}. It is composed of a deep convolutional network and a Long Short-Term Memory(LSTM), where the deep CNN is directly applied for feature extraction, and a bidirectional LSTM network is applied for sequence labeling. DenseNet\cite{huang2017densely} is a highly efficient convolutional neural network. Because of its low parameter number and fast inference time, DenseNet is widely used. Our method is also a segmentation-free approach based on the framework proposed by \cite{huang2017densely}, where DenseNet is applied for feature extraction.

Data generation is used in license plate recognition to improve the accuracy of recognition. The labeled license plates generated by CycleGAN as a pre-training data set for the recognition network are used in \cite{wang2017adversarial}, and the model is fine-tuned with the real license plate data set. This data generation method can significantly improve the recognition accuracy. License plate detection and recognition is combined in \cite{li2018toward}, and it finally improves the recognition speed and recognition accuracy of the system.

\subsection{Generative Adversarial Networks}
Generative adversarial networks(GANs) \cite{goodfellow2014generative,radford2015unsupervised} train a generator and discriminator alternatively. The output of the discriminator acts as a generator's loss function. Zhu et al. \cite{zhu2017unpaired} propose Cycle-Consistent Adversarial Networks(CycleGAN), which learns the mapping relationship from one domain to another and is mainly used for the style conversion of pictures. Wasserstein GANs(WGAN) \cite{arjovsky2017wasserstein} are proposed to improve the stability of GAN training. Applying Wasserstein loss to CycleGAN, creating CycleWGAN, also improves its training stability in \cite{wang2017adversarial}. Gradient penalties in WGAN(WGAN-GP) \cite{gulrajani2017improved} are proposed to solve the WGAN generator weight distribution problem.



\subsection{Data Generation For Training}
A large number of real labeled images are often difficult to obtain, so the role of data generation is very significant \cite{shrivastava2017learning}. The synthesized images are used to train scene text detection networks \cite{gupta2016synthetic} and recognition networks \cite{jaderberg2014synthetic}. The generated data is shown to improve the performance of person detection \cite{yu2010improving}, font recognition \cite{wang2015deepfont}, and semantic segmentation \cite{ros2016synthia}. However, when the difference between the generated data and the real data is very large, the performance is poor when applied to a real scene. Therefore, \cite{wang2017adversarial} applies CycleGAN to convert the style of license plate generated by the script into a real license plate, which can greatly reduce the gap between the generated image and the real image. We apply data generation and data augmentation methods at the same time, and use the data generated by different methods directly as the training set for recognition network. Therefore we need very little real data.

\section{License plate recognition based on data generation and augmentation.}
In this section, the pipeline of the proposed method is described. We train the GAN model using synthetic images and real images simultaneously. We then use the generated images to train a model modified from DenseNet.
\subsection{CycleGAN}
CycleGAN \cite{zhu2017unpaired} learns to translate an image from a source domain X to a target domain Y in the absence of paired examples. Our goal is to train a mapping relationship G between the script license plate domain X and the real license plate domain Y. CycleGAN contains two mapping functions $G:X\rightarrow Y$ and $Y\rightarrow X$, and associated adversarial discriminators D$_Y$, D$_X$.

The techniques proposed in WGAN \cite{arjovsky2017wasserstein} are applied in CycleGAN, and CycleWGAN is proposed in \cite{wang2017adversarial}. WGAN points out why the traditional GAN is difficult to converge and improve during training, which greatly reduces the training difficulty and speeds up the convergence. There are two main improvements: the first one is to remove the log from the loss function, and the second is to perform weight clipping after each iteration to update the weight, and limit the weight to a range (eg, the limit range is [-0.1, +0.1]. Outside weights are trimmed to -0.1 or +0.1). CycleWGAN solves the problem of training instability and collapse mode, which makes the result more diverse.

We apply the techniques in WGAN-GP \cite{gulrajani2017improved} to CycleWGAN and propose the CycleWGAN-GP. WGAN-GP also proposes an improvement plan based on WGAN. WGAN reduces the training difficulty of GAN, but it is still difficult to converge in some conditions, and the generated pictures are worse than DCGAN. WGAN-GP applies gradient penalty, and solves the above problem along with the problems of vanishing gradient and exploding gradient during training. It also converges faster than CycleWGAN and produces higher quality pictures.

We apply a CycleGAN equipped with WGAN and WGAN-GP techniques to train the mapping relationship between the fake license plate and the real license plate. First of all, we apply OpenCV scripts to generate synthetic license plates as a source domain X, and then choose real license plates without labels as a target domain Y. Before the training of CycleWGAN-GP, these license plates are randomly cropped and randomly flipped horizontally or vertically.


\subsection{Recognition network design}
DenseNet is a densely connected convolutional neural network. In this network, there is a direct connection between any two layers. The input of each layer of the network is the union of the output of all previous layers, and the feature map learned by this layer is also directly transmitted to all subsequent layers. DenseNet allows the input of l$^{th}$ Layer to directly affect all subsequent layers.
Its output is:
\begin{equation}
x_l=H_l([x_0,x_1,...,x_{l-1}]) \label{equ:dense}
\end{equation}
where $H_l(\cdot)$ refers to a composite function of three consecutive operations: batch normalization (BN) \cite{Ioffe2015Batch}, followed by a rectified linear unit (ReLU), and a $3\times 3$ convolution (Conv).
Additionally, since each layer contains the output information of all previous layers, it only needs a few feature maps, so the number of parameter of DenseNet is greatly reduced compared to other models.

\begin{table}[t]
\centering
\caption{Construction of recognition network. The output size represents w$\times$h$\times$c. Note that each ¡°conv¡± layer shown in the table corresponds the sequence BN-ReLU-Conv}
\label{tab:construction}
\begin{tabular}{|c|c|c|}
\hline
Layers                               & Output Size                  &  Recognition Network                   \\ \hline
Input                                & 136$\times$36$\times$1       &                      \\ \hline
Convolution                          & 68$\times$18$\times$64       & 5$\times$5 conv, stride 2         \\ \hline
Dense Block(1)                       & 68$\times$18$\times$128      & {[} 3$\times$3 conv {]} $\times$ 8       \\ \hline
\multirow{2}{*}{Transition Layer(1)} & 68$\times$18$\times$128      & 1$\times$1 conv                   \\ \cline{2-3}
                                     & 34$\times$9$\times$128       & 2$\times$2 average pool, stride 2 \\ \hline
Dense Block(2)                       & 34$\times$9$\times$192       & {[} 3$\times$3 conv {]} $\times$ 8       \\ \hline
\multirow{2}{*}{Transition Layer(2)} & 34$\times$9$\times$128       & 1$\times$1 conv                   \\ \cline{2-3}
                                     & 17$\times$4$\times$128       & 2$\times$2 average pool, stride 2 \\ \hline
Dense Block(3)                       & 17$\times$4$\times$192       & {[} 3$\times$3 conv {]} $\times$ 8       \\ \hline

\end{tabular}
\end{table}

Our network structure is shown in Table \ref{tab:construction}, which is different from the network structure of \cite{huang2017densely}, because the input license plate image is smaller and is a gray scale image of 136$\times$36, so the network only has 3 dense blocks. The transition layers used in our network consist of a batch normalization layer and an 1$\times$1 convolutional layer followed by a 2$\times$2 average pooling layer. A 1$\times$1 convolution can be introduced
as bottleneck layer before each 3$\times$3 convolution to reduce the number of input feature-maps. To improve model compactness, we reduce the number of feature-maps from 192 to 128 at transition layers 2.

The last DenseNet layer is followed by a fully-connected layer with 68 neurons for the 68 classes of label, including 31 Chinese characters, 26 letters, 10 digits and ¡°blank¡±. We train the networks with stochastic gradient descent (SGD). The labelling loss is derived using Connectionist Temporal Classification (CTC) \cite{graves2006connectionist}. The optimization algorithm Adam \cite{kingma2014adam} is then applied, as it converges quickly and does not require a complicated learning rate schedule. Another advantage of using the modified DenseNet network is that it does not require the Long Short-Term Memory(LSTM) networks. The use of LSTM complicates the solution and increases computational cost.

\section{Experiment}
In this section, we conduct experiments to verify the effectiveness of the proposed methods. Our network is implemented
capitalizing keras. The experiments are trained on a NVIDIA Tesla P40 with 24GB memory and are tested on a NVIDIA GTX745 GPU with 4GB memory.
\subsection{Dataset}
The image in the Dataset-1 \cite{wang2017adversarial} are captured from a wide variety of real traffic monitoring scenes under various viewpoints, blurring and illumination. Dataset-1 contains a training set of 203,774 plates and a test set of 9,986 plates. The first character of Chinese license plates is a Chinese character which represents the province. While there are 31 abbreviations for all of the provinces, Dataset-1 contains 30 classes of them.

The second data set is the application-oriented license plate (AOLP) \cite{hsu2013application} benchmark database, which has 2049 images of Taiwan license plates. This database is categorized into three subsets: access control (AC) with 681 samples, traffic law enforcement (LE) with 757 samples, and road patrol (RP) with 611 samples.

\subsection{Implementation Details}
\subsubsection{Network}
The recognition network is shown in Table \ref{tab:construction}. We implement it with Keras. The images are resized to 136$\times$36 and converted to gray scale and then fed to the recognition network. We change the last layer of fully connected layers to 68 neurons according to the 68 classes of characters-33 Chinese characters, 24 letters, 10 digits and "blank". We train the networks with SGD and learning rate of 0.0001. The labelling loss is derived using CTC. We set the training batch size as 256 and predicting size as 1.
\subsubsection{Evaluation Criterion}
In this work, we evaluate the model's performance in terms of recognition accuracy and character recognition accuracy, which is similar to Wang et al.\cite{wang2017adversarial}. Recognition accuracy is defined as :
\begin{equation}
RA = \frac{Number\ of\ correctly\ recognized\ license\ plates}{Number\ of\ all\ license\ plates}
\end{equation}
Character recognition accuracy is defined as:
\begin{equation}
CRA = \frac{Number\ of\ correctly\ recognized\ characters}{Number\ of\ all\ characters}
\end{equation}
\begin{figure*}[t]
    \centering
        \subfigure[]{\label{fig:script}
        \includegraphics[width=0.5\columnwidth]{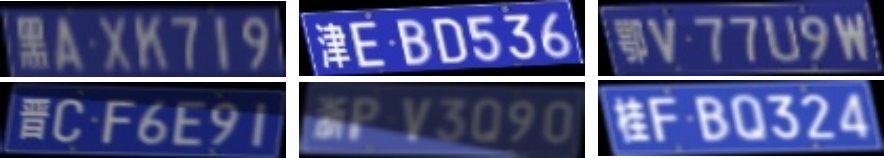}
        }
        \subfigure[]{\label{fig:WGAN}
        \includegraphics[width=0.5\columnwidth]{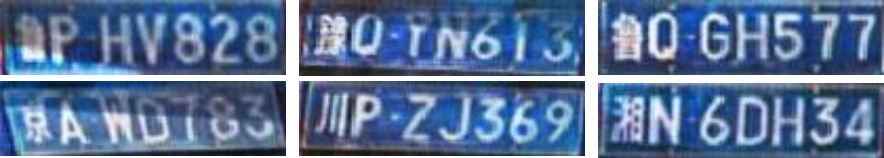}
        }
        \subfigure[]{\label{fig:WGAN-GP}
        \includegraphics[width=0.5\columnwidth]{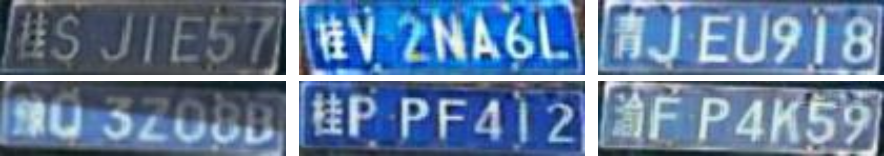}
        }
    \caption{Three data generation methods (a)Examples of license plates generated by OpenCV scripts. (b)Examples of license plates generated by CycleWGAN. (c)Examples of license plates generated by CycleWGAN-GP. }
    \label{fig:tradeoff}
    \end{figure*}
\subsubsection{GAN Training and Testing}
Three data generation methods are shown in Fig. \ref{fig:tradeoff}. To train CycleWGAN, first we use the OpenCV scripts to generate 1000 blue fake license plates as a source domain X, and then select 1000 real blue license plates from Dataset-1 as a target domain Y. We train the CycleWGAN model with these fake license plates and real license plates. The training real plates do not require character labels. All the images are resized to 143$\times$143, cropped to 128$\times$128 and randomly flipped for data augmentation. We use Adam with $L_{1}=0.9$, $L_{2}=0.999$ and learning rate of 0.0001. We stop training after 300,000 steps and save the model. When testing, first we use the OpenCV scripts to generate 40,000 blue fake license plates, and then we apply the last checkpoint to generate 40,000 license plates. The same goes for CycleWGAN-GP. Finally we get 80,000 blue license plates generated by CycleWGAN and CycleWGAN-GP.

\subsubsection{Data Augmentation}
\begin{figure*}[t]
    \centering
        \subfigure[]{\label{fig:a1}
        \includegraphics[width=0.40\columnwidth]{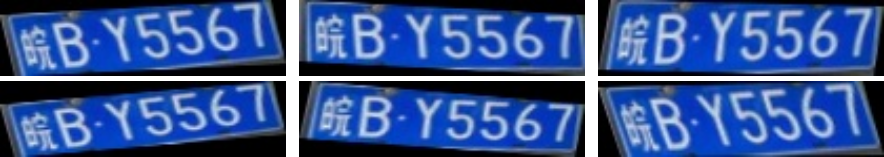}
        }
        \subfigure[]{\label{fig:a4}
        \includegraphics[width=0.40\columnwidth]{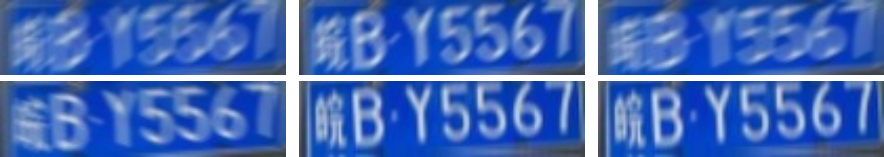}
        }
        \subfigure[]{\label{fig:a5}
        \includegraphics[width=0.40\columnwidth]{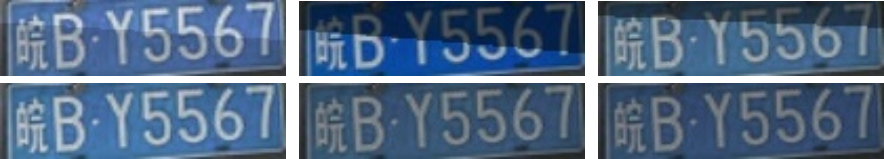}
        }
        \subfigure[]{\label{fig:a6}
        \includegraphics[width=0.40\columnwidth]{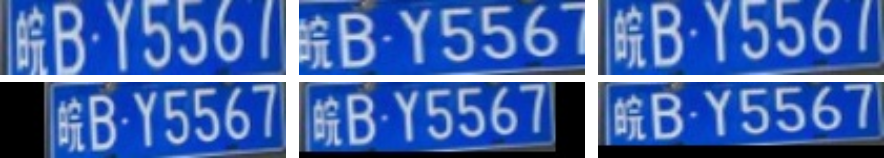}
        }
        \subfigure[]{\label{fig:a7}
        \includegraphics[width=0.40\columnwidth]{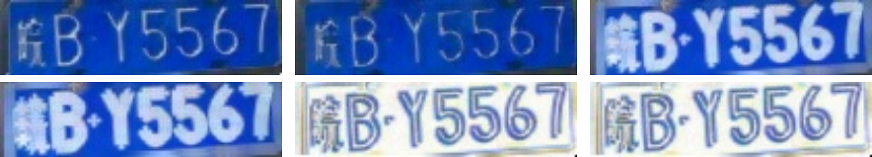}
        }
        \subfigure[]{\label{fig:a8}
        \includegraphics[width=0.40\columnwidth]{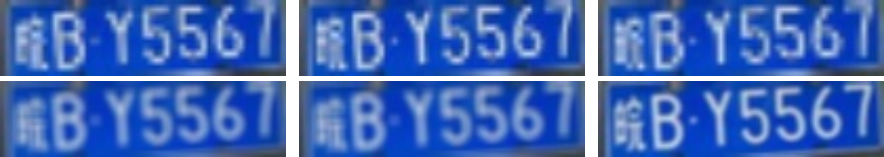}
        }
    \caption{Six data augmentation methods(a)Affine transformation (b)Motion blur (c)Uneven light (d)Stretching transformation (e)Erosion and dilation (f)Down sampling and gaussian noise.}
    \label{fig:augm}
    \end{figure*}

The six data augmentation methods are proposed in order to increase the training data of the recognition network. The data was augmented through affine transformation, motion blurring, uneven lighting, stretching, erosion and dilation, downsampling and the application of gaussian noise. Examples of these transformations are shown in Fig. \ref{fig:augm}. A real license plate image randomly passes through the six data augmentation methods, allowing for the creation of much more training data. First, we select a small number of labeled real license plates from Dataset-1, such as 300. And then using data augmentation methods in Fig. \ref{fig:augm}, we generate 80,000 augmented license plates with these selected real license plates.

\subsubsection{Mixed Training Data}
Our mixed training data consists of four parts, including 40,000 license plates generated by OpenCV scripts, 40,000 license plates generated by CycleWGAN, 40,000 license plates generated by CycleWGAN-GP, and 80,000 license plates augmented from a small number of labeled real license plates. All 200,000 training images are generated with license plate character labels. The license plates that need manual labeling are only selected from Dataset-1. After converting the training data to gray scale, 400,000 more training images are obtained by flipping pixels in order to simulate gray images of yellow and green license plates. Then, these images are fed to the recognition network modified from DenseNet.

\subsection{Performance Evaluation on Dataset-1}

\begin{table}
\caption{The accuracy of other methods compared with the proposed method on Dataset-1. Recognition accuracy (\%) and character recognition accuracy (\%) are listed.}\label{tab:dataset1}
\centering
\begin{tabular}{|p{3cm}<{\centering}|p{2cm}<{\centering}|p{2cm}<{\centering}|p{2cm}<{\centering}|}
\hline
Method & Training Data & RA & CRA \\
\hline
\multirow{3}*{baseline} & 9000 & 96.1 & 98.9 \\
		~ & 50000 & 96.7 & 99.1 \\
        ~ & 200000 & 97.6 & 99.5 \\
\hline
\multirow{3}*{ours} &300 & 97.5 & 99.3\\
        ~ & 700 & 98.2 & 99.5\\
        ~ & 3333 & 98.6 & 99.8\\
        ~ & 4750 & 99.0 & 99.9\\
        ~ & 6000 & 99.0 & 99.9\\
\hline
\end{tabular}
\end{table}

With the above methods, our mixed training data is generated from 300, 700, 3,333, 4,750 and 6,000 real license plates selected from Dataset-1 training set respectively. Our baseline is the \cite{wang2017adversarial} using the license plate images generated by the CycleWGAN pre-training recognition network, and then using 9,000, 50,000 and 200,000 real labeled license plate images in a fine-tuning model. From the results in Table \ref{tab:dataset1}, it is concluded that when data generation, data augmentation and DenseNet are used, we only need 300 real labeled license plates to achieve the effect of 200,000 real license plates. In the same way, when the number of real license plates reaches 4,750, the final recognition accuracy has reached 99.0\%, an increase of 1.4\%. When the number of real license plate images exceeds 4,750, license plate recognition accuracy and character recognition accuracy are not improving. We conjecture that 4,750 real images contain enough information to recognize most of the license plates. Thus, by increasing the number of real license plates, the total amount of information after data augmentation will not change, and the recognition accuracy will not increase any further.

\subsection{Performance Evaluation on Data Generation}
In order to evaluate the effect of the data generated by different methods, we train the models using synthetic data generated by script, CycleGAN, CycleWGAN, and CycleWGAN-GP respectively. The results are shown in Table \ref{tab:augmentation}. When we only use the data set generated by script for training, the recognition accuracy on the test set of Dataset-1 is 42.2\%. As shown in Fig. \ref{fig:script}, our synthetic license plates generated by script also contain noise such as low light, low resolution, motion blur. The CycleGAN images achieve a recognition accuracy of 51.2\%. Accuracy is not much improved because of  the instability and lack of diversity in CycleGAN training. As shown in Fig. \ref{fig:WGAN} and Fig. \ref{fig:WGAN-GP}, the CycleWGAN and CycleWGAN-GP images display more various styles and colors, and part of them can not really distinguish from real images. The CycleWGAN and CycleWGAN-GP images achieve a recognition accuracy of 62.5\% and 64.5\% respectively. We also compare the impact of data generation and data augmentation on accuracy. When the number of real license plates is 3333, the recognition accuracy of the augmented data on Dataset-1 is 97.9\%, far exceeding the recognition accuracy of the generated data.

\begin{table}[t]
\centering
\caption{Single data augmentation recognition results with 3333 real images compare with \cite{wang2017adversarial}. Recognition accuracy (\%), character recognition accuracy (\%) are shown. "CRA-C" is the recognition accuracy (\%) of the Chinese characters of the first character, and "CRA-NC" is the recognition accuracy (\%) of the letters and numbers of the last six characters.}\centering\label{tab:augmentation}
\label{my-label}
\begin{tabular}{|p{2.4cm}<{\centering}|p{1cm}<{\centering}|p{1cm}<{\centering}|p{1.1cm}<{\centering}|p{1.1cm}<{\centering}|p{1cm}<{\centering}|p{1cm}<{\centering}|p{1.1cm}<{\centering}|p{1.35cm}<{\centering}|}
\hline
             & \multicolumn{4}{c}{ours}                     \vline     & \multicolumn{4}{c}{baseline}          \vline    \\
\hline\
method       & RA   & CRA  & CRA-C & \multicolumn{1}{c|}{CRA-NC} & \multicolumn{1}{c|}{RA} & CRA  & CRA-C & CRA-NC \\
\hline
Script       & 42.2 & 80.3 & 43.8  & 90.8                        & 4.4                     & 30.0 & 20.0  & 31.7   \\
CycleGAN     & 51.2 & 87.6 & 51.2  & 93.1                        & 34.6                    & 82.8 & 41.3  & 89.8   \\
CycleWGAN    & 62.5 & 92.5 & 66.8  & 96.8                        & 61.3                    & 90.6 & 66.2  & 94.8   \\
CycleWGAN-gp & 64.5 & 93.7 & 65.2  & 98.4                        & -                       & -    & -     & -      \\
Augmentation & 97.9 & 99.1 & 99.2  & 99.7                        & -                       & -    & -     & -       \\
\hline
\end{tabular}
\end{table}

In order to understand how much number of real license plates improves recognition accuracy, we compare data augmentation results from 60 to 6000 real license plates. The result in Table \ref{tab:numdata} shows that the greater the number of real license plates, the higher the recognition accuracy obtained. Up to 4750, the highest recognition accuracy of the Dataset-1 is 99.0\%. Even if the number of real license plates is increased from 4750, the result is no longer improved.

In order to understand the impact of GAN on recognition accuracy, we did some additional comparative experiments. It can also be seen in the Table \ref{tab:numdata} that training data composed of data augmentation and data generation can get better results than training data composed of only data augmentation. The conclusion is that the recognition accuracy of augmented data can be improved with data generation. In addition, the fewer real license plates, the more recognition accuracy increases contributed from generated data.
\begin{table}[t]
\caption{The effect of the number of plates and data generation on the results. Recognition accuracy (\%), character recognition accuracy (\%) are shown. RA(A) indicates the recognition accuracy of the data augmentation. CRA(A) indicates the character recognition accuracy of the data augmentation. RA(A+G) indicates the recognition accuracy of the training data composed of data augmentation and data generation. CRA(A+G) indicates the character recognition accuracy of the mixed training data composed of data augmentation and data generation.}\centering\label{tab:numdata}
\begin{tabular}{|p{3cm}<{\centering}|p{2cm}<{\centering}|p{2cm}<{\centering}|p{2cm}<{\centering}|p{2cm}<{\centering}|}
\hline
Training Data &  RA(A) & RA(A+G) & CRA(A) & CRA(A+G)\\
\hline
60 &  47.5 & 79.3 & 88.3 & 94.4\\
150 &  83.8 & 93.8 & 96.6 & 98.7\\
200 &  92.4 & 96.7 & 98.4 & 98.7\\
300 &  96.1 & 97.5 & 98.7 & 99.3\\
700 &  97.1 & 98.2 & 98.9 & 99.5\\
3333 &  97.9 & 98.6 & 99.2 & 99.8\\
4750 &  98.8 & 99.0 & 99.8 & 99.9\\
6000 &  98.9 & 99.0 & 99.8 & 99.9\\
\hline
\end{tabular}
\end{table}

\subsection{Performance Evaluation on AOLP}
For the application-oriented license plate(AOLP) dataset, the experiments are carried out by using license plates from different sub-datasets for training and test. This data set is divided into three sub-datasets: access control (AC), traffic law enforcement (LE), and road patrol (RP). For example, in Table \ref{tab:alop}, we use the license plates from the LE and RP sub-datasets to train the DenseNet, and test its performance on the AC sub-dataset. Similarly, AC and RP are used for training and LE for test, and so on. Since there is no AOLP license plate font, only the data augmentation methods are used, without script and GAN generated license plates. In Table \ref{tab:alop}, through data augmentation and DenseNet, our method achieves the highest recognition accuracy on the AOLP dataset.

\begin{table}[t]
\centering
\caption{The accuracy of other methods compared with the proposed method on dataset AOLP. Recognition accuracy (\%), character recognition accuracy (\%) are shown. AOLP is categorized into three subsets: access control (AC) with 681 samples, traffic law enforcement (LE) with 757 samples, and road patrol (RP) with 611 samples}
\label{tab:alop}
\begin{tabular}{|p{2cm}<{\centering}|p{1.2cm}<{\centering}|p{1.2cm}<{\centering}|p{1.2cm}<{\centering}|p{1.2cm}<{\centering}|p{1.2cm}<{\centering}|p{1.2cm}<{\centering}|}
\hline
           & \multicolumn{2}{c}{AC(\%)}  \vline     & \multicolumn{2}{c}{LE(\%)}  \vline    & \multicolumn{2}{c}{RP(\%)} \vline\\
\hline
method     & RA    & CRA & RA & CRA   & RA & CRA \\
\hline
Hsu et al.\cite{hsu2013application} & -     & 96                       & -                       & 94    & -             & 95         \\
Li et al.\cite{li2016reading}  & 94.85 & -                        & 94.19                   & -     & 88.38         & -          \\
Li et al.\cite{li2018toward}  & 95.71 & -                        & 97.21                   & -     & 84.60         & -          \\
ours       & \textbf{96.61} & \textbf{99.08}            & \textbf{97.80}           & \textbf{99.65} & \textbf{91.00}         & \textbf{97.22}        \\
\hline
\end{tabular}
\end{table}


\section{Conclusion}
In this paper, we have investigated how many real labeled license plates are needed to train the license plate recognition system. We have proposed three data generation methods and six data augmentation methods in order to fully obtain all the information in a small number of images. The experimental results show that the proposed method only requires 300 real labeled license plates to achieve the effect achieved by 200,000 real license plates. The result shows that the greater the number of real license plates, the higher the recognition accuracy obtained. Up to 4750, the highest recognition accuracy of the Dataset-1 is 99.0\%. Even if the number of real license plates is increased furthermore, the result is no longer improved. Additionally, training data composed of both augmented and generated data can achieve better results than training data composed of only augmented data. Furthermore, the fewer real license plates, the more recognition accuracy increases contributed from generated data.

%
%
%
\bibliographystyle{splncs04}
\bibliography{myreference}

\end{document}